%% file: main.tex
\documentclass{article} 

\usepackage[accepted]{icml2020}

\usepackage{amsmath}
\usepackage{amssymb}
\usepackage{amsthm}
\usepackage{chngcntr}   
\usepackage{dsfont}
\usepackage{enumitem}   
\usepackage{nccmath}    
\usepackage{mathtools}  
\usepackage{mleftright} 
\usepackage{etoolbox}   

\usepackage{hyperref}
\usepackage{url}

\usepackage{booktabs}
\usepackage[olditem, oldenum]{paralist}
\usepackage{multirow}
\usepackage{nicefrac}
\usepackage{prettyref}
\usepackage{siunitx}
\usepackage{tabularx}
\usepackage{xspace}

\usepackage{pgfplots}
\usepackage{tikz}

\pgfplotsset{compat=1.14}
\usepgfplotslibrary{colorbrewer}
\usepgfplotslibrary{groupplots}

\usepackage{etoolbox}
\robustify\bfseries
\robustify\itshape

\renewcommand{\comment}[1]{}

\bibpunct{\nolinebreak{}(}{)}{,}{a}{,}{,}

\DeclareMathOperator{\im}    {im}

\newcommand{\E}        {\mathds{E}}
\renewcommand{\natural}{\mathds{N}\xspace}
\newcommand{\real}     {\mathds{R}\xspace}

\newcommand{\attention}{a}

\renewcommand{\th}{\textsuperscript{\textup{th}}\xspace}

\usetikzlibrary{arrows}
\usetikzlibrary{calc}
\usetikzlibrary{chains}
\usetikzlibrary{positioning}

\newcommand{\methodname}     {\textsc{SeFT}\xspace}
\newcommand{\methodnamelong} {\textbf{Se}t \textbf{F}unctions for \textbf{T}ime
Series\xspace}

\newcommand{\dataset}[1]{\texttt{#1}}
\newcommand{\method}[1]{\textsc{#1}}
\newcommand{\OOM}{OOM}

\newtheorem{definition}{Definition}
\newrefformat{fig}{Figure~\ref{#1}}
\newrefformat{appx}{Appendix~\ref{#1}}
\newrefformat{tab}{Table~\ref{#1}}
\newrefformat{eq}{Equation~\ref{#1}}

\DeclareSIUnit\epoch{epoch}

\begin{document}

\twocolumn[
\icmltitle{Set Functions for Time Series}

\begin{icmlauthorlist}
  \icmlauthor{Max Horn}{ethz,sib}
  \icmlauthor{Michael Moor}{ethz,sib}
  \icmlauthor{Christian Bock}{ethz,sib}
  \icmlauthor{Bastian Rieck}{ethz,sib}
  \icmlauthor{Karsten Borgwardt}{ethz,sib}
\end{icmlauthorlist}

\icmlaffiliation{ethz}{Department of Biosystems Science and Engineering, ETH Zurich, 4058 Basel, Switzerland}
\icmlaffiliation{sib}{SIB Swiss Institute of Bioinformatics, Switzerland}

\icmlcorrespondingauthor{Karsten Borgwardt}{karsten.borgwardt@bsse.ethz.ch}

\icmlkeywords{Machine Learning, ICML}

\vskip 0.3in
]
\printAffiliationsAndNotice{}

\begin{abstract}
Despite the eminent successes of deep neural networks, many
architectures are often hard to transfer to irregularly-sampled and
asynchronous time series that commonly occur in real-world datasets,
especially in healthcare applications. This paper proposes a novel
approach for classifying irregularly-sampled time series with unaligned
measurements, focusing on high scalability and data efficiency. Our
method SeFT~(\textbf{S}et \textbf{F}unctions for \textbf{T}ime Series)
is based on recent advances
in differentiable set function learning, extremely parallelizable with
a beneficial memory footprint, thus scaling well to large datasets of
long time series and online monitoring scenarios. Furthermore, our
approach permits quantifying per-observation contributions to the
classification outcome. We extensively compare our method with existing
algorithms on multiple healthcare time series datasets and demonstrate
that it performs competitively whilst significantly reducing runtime.
\end{abstract}

\section{Introduction}

With the increasing digitalization, measurements over extensive time periods
are becoming ubiquitous.
Nevertheless, in many application domains, such as healthcare~\citep{yadav2018mining},
measurements might not necessarily be
observed at a regular rate or could be misaligned. Moreover, the presence or
absence of a measurement and its observation frequency may carry information
of its own~\citep{little2014statistical}, such that imputing the missing values
is not always desired.

While some algorithms can be readily applied to datasets with varying
lengths, these methods usually assume \emph{regular} sampling of the data and/or
require the measurements across modalities to be aligned/synchronized,
preventing their application to the aforementioned settings.
By contrast, existing approaches, in particular in clinical applications, for
\emph{unaligned} measurements, typically rely on imputation
to obtain a regularly-sampled version of a dataset for
classification~\citep{desautels2016prediction, Moor19}.
Learning a suitable imputation scheme, however, requires understanding the
underlying dynamics of a system; this task is significantly more complicated
and \emph{not} necessarily required when classification or pattern
detection is the main goal.
Furthermore, even though a decoupled imputation scheme followed by
classification is generally more scalable, it may lose information~(in
terms of ``missingness patterns'') that could be crucial for prediction
tasks.
The fact that decoupled schemes perform worse than methods
that are trained end-to-end was empirically
demonstrated by \citet{li2016scalable}.
Approaches that jointly optimize both tasks add a large
computational overhead, thus suffering from poor scalability or high
memory requirements.

Our method is motivated by the understanding that, while RNNs and
similar architectures are well suited for capturing and modelling the
dynamics of a time series and thus excel at tasks such as forecasting,
retaining the order of an input sequence can even be a disadvantage in
some scenarios~\citep{vinyals2015order}.
We show that by relaxing the condition that a sequence must be processed
in order, we can naturally derive an architecture that \emph{directly}
accounts for
\begin{inparaenum}[(i)]
  \item irregular sampling, and
  \item unsynchronized measurements.
\end{inparaenum}
Our method \methodname: \methodnamelong, extends recent advances in set
function learning to irregular sampled time series classification tasks,
yields favourable classification performance, is highly scalable, and improves
over current approaches by almost an order of magnitude in terms of
runtime.
With \methodname, we propose to rephrase the problem of classifying time
series as classifying a set of observations. We show how set
functions can be used to create classifiers that are applicable
to unaligned and irregularly-sampled time series, leading to favourable
performance in classification tasks. Next to being highly
parallelizable, thus permitting ready extensions to online monitoring
setups with thousands of patients, our method also yields
importance values for each observation and each modality.  This makes it
possible to \emph{interpret} predictions, providing much-needed
insights into the decision made by the model.

\section{Related Work}

This paper focuses on classifying time series with irregular sampling
and potentially unaligned measurements. We briefly discuss recent
work in this field; all approaches can be broadly grouped into the
following three categories.

\begin{enumerate}[nosep, parsep=0.5em, labelindent=0em, itemsep=0em,
  align=left, labelwidth=*, leftmargin=0em]
\item[\textbf{Irregular sampling as missing data}:]
While the problem of supervised classification in the presence of missing data
is closely related to irregular sampling on time series, there are some core
differences. Missing data is usually defined with respect to a number of
features that could be observed, whereas time series themselves can have different
lengths and a ``typical'' number of observed values might not exist.
Generally, an irregularly-sampled time series can be converted into a
missing data problem by discretizing the time axis into non-overlapping
intervals, and declaring intervals in which no data was sampled as
missing~\citep{Bahadori19}.
This approach is followed by \citet{marlin2012unsupervised}, who used a Gaussian
Mixture Model for semi-supervised clustering on electronic health
records.
Similarly, \citet{lipton2016directly} discretize the time series into
intervals, aggregate multiple measurements within an interval, and add
missingness indicators to the input of a Recurrent Neural Network.
By contrast, \citet{che2018recurrent} present several variants of the Gated
Recurrent Unit~(GRU) combined with imputation schemes. Most prominently, the
GRU-model was extended to include a decay term~(GRU-D), such that the last
observed value is decayed to the empirical mean of the time series via a
learnable decay term. While these approaches are applicable to
irregularly-sampled data, they either rely on imputation schemes or
empirical global estimates on the data distribution~(our method, by
contrast, requires neither), without directly
exploiting the global structure of the time series.

\item[\textbf{Frameworks supporting irregular sampling}:]
Some frameworks support missing data. For example,
\citet{lu2008reproducing} directly defined a kernel on
irregularly-sampled time series, permitting subsequent classification
and regression with kernel-based classifiers or regression schemes.
Furthermore, Gaussian Processes~\citep{williams2006gaussian} constitute
a common probabilistic model for time series; they directly permit
modelling of continuous time data using mean and covariance functions.
Along these lines, \citet{li2015classification} derived
a kernel on Gaussian Process Posteriors, allowing the comparison and
classification of irregularly-sampled time series using
kernel-based classifiers.
Nevertheless, all of these approaches still rely on separate
tuning/training of the imputation method and the classifier so that
structures supporting the classification could be potentially missed in
the imputation step.
An emerging line of research employs \emph{Hawkes
processes}~\citep{Hawkes71, Liniger09}, i.e.\ a specific class of self-exciting
point processes, for time series modelling and
forecasting~\citep{Mei17, Yang17, Xiao17}.
While Hawkes processes exhibit extraordinary performance in these
domains, there is no standardised way of using them for classification.
Previous work~\citep{Lukasik16} trains \emph{multiple} Hawkes
processes~(one for each label) and classifies a time series by assigning
it the label that maximises the respective likelihood function. Since
this approach does not scale to our datasets, we were unable to perform
a fair comparison.  We conjecture that further research will be required
to make Hawkes processes applicable to general time series
classification scenarios.

\item[\textbf{End-to-end learning of imputation schemes}:]
Methods of this type are composed of two modules with separate
responsibilities, namely an imputation scheme and a classifier, where
both components are trained in a discriminative manner and end-to-end using
gradient-based training.
Recently, \citet{li2016scalable} proposed the Gaussian Process
Adapters~(GP Adapters) framework, where the parameters of a Gaussian
Process Kernel are trained alongside a classifier. 
The Gaussian Process gives rise to a fixed-size representation of the
irregularly-sampled time series, making it possible to  apply \emph{any}
differentiable classification architecture. This approach was further
extended to multivariate time series by \citet{futoma2017learning} using
Multi-task Gaussian Processes~(MGPs) \citep{bonilla2008multi}, which
allow correlations between the imputed channels.
Moreover, \citet{futoma2017learning} made the approach more compatible
with time series of different lengths by applying a Long Short Term
Memory (LSTM)~\citep{hochreiter1997long} classifier.
Motivated by the limited scalability of approaches based on GP Adapters,
\citet{shukla2018interpolationprediction} suggest an alternative
imputation scheme, the \emph{interpolation prediction networks}.
It applies multiple semi-parametric
interpolation schemes to obtain a regularly-sampled time series
representation. The parameters of the interpolation network are trained
with the classifier in an end-to-end setup.

\end{enumerate}

\section{Proposed Method}

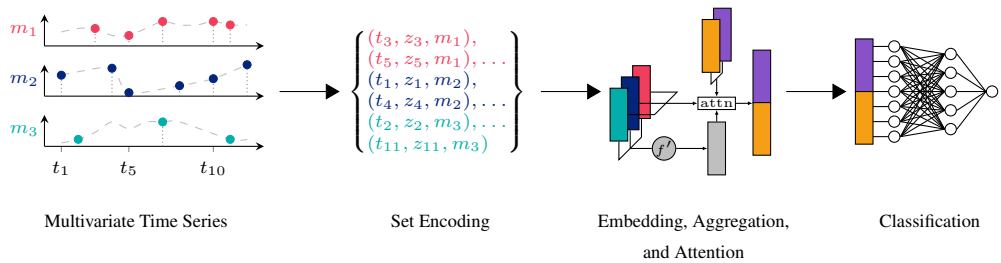
\begin{figure*}[tb]
	\centering
	\begin{tikzpicture}[scale=1.00, start chain=going right, node distance = 8mm, ]
	\tikzset{
		block/.style = {%
			draw,
			on chain,
			on grid,
			rounded corners,
			align          = center,
			fill           = white,
			draw           = none,
			inner sep      = 3pt,
			shape          = rectangle,
			minimum height = 60pt,
		},
		line/.style = {%
			black,
			>=triangle 60,
			draw,
			->,
		}
	}
	\def\heightBox{3cm}
	\node[block, label=below:{\scriptsize Multivariate Time Series},minimum height=\heightBox] (MultiTime) {
		\centering
		\includegraphics[]{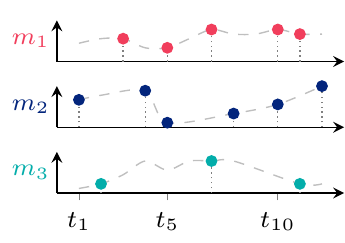}
	};
	\node[block, label=below:{\scriptsize Set Encoding},minimum height=\heightBox] (Sets) {
		\centering
		\includegraphics[]{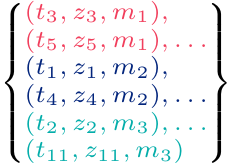}
	};
	\node[block, label={[align=center]below:\scriptsize Embedding, Aggregation, \\ \scriptsize and Attention},minimum height=\heightBox] (Embedding) {
		\centering
		\includegraphics[]{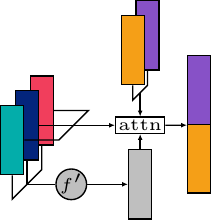}
	};
	\node[block, label=below:{\scriptsize Classification},minimum height=\heightBox] (Final) {
		\centering
		\includegraphics[]{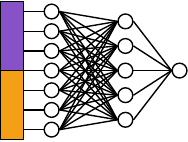}
	};
	
	\draw[->, line] (MultiTime) -- (Sets);
	\draw[->, line] (Sets) -- (Embedding);
	\draw[->, line] (Embedding) -- (Final);
	
	\end{tikzpicture}
	\caption{%
      Schematic overview of \methodname's architecture. The first
      panel exemplifies a potential input, namely a multivariate time
      series, consisting of $3$ modalities $m_1, m_2, m_3$. We treat the
      $j$\th observation as a tuple $\left(t_j, z_j, m_j \right)$,
      comprising of a time $t_j$, a value $z_j$, and a modality indicator
      $m_j$.
      All observations are summarized as a set 
      of such tuples. The elements of each set are summarized using a set function $f'$.
      Conditional on the summarized representation and the individual set elements
      an attention mechanism (as described in Section~{\ref{sec:attention}}) is applied 
      to learn the importance of the individual observations. Respective
      query vectors for \num{2} attentions head are illustrated in purple and orange blocks.
      The results of each attention head are then concatenated and used as the input for the
      final classification layers.
	}
	\label{fig:Workflow}
\end{figure*}

Our paper focuses on the problem of time series classification of irregularly
sampled and unaligned time series. We first define the required terms before
describing our model. 

\subsection{Notation \& Requirements}

\begin{definition}[Time series]
  We describe a time series of an instance $i$ as a set $\mathcal{S}_i$
  of $M := |\mathcal{S}_i|$ observations
  $s_j$ such that $\mathcal{S}_i := \left\{s_1, \dots, s_M \right\} $.
  We assume that each observation $s_j$ is represented as a tuple
  $\left(t_j, z_j, m_j \right)$, consisting of a time value $t_j \in \real^{+}$,
  an observed value $z_j \in \real$,
  and a modality indicator $m_j \in \{1 \dots D\}$, where $D$ represents the dimensionality
  of the time series.
  We write $\Omega \subseteq \real^{+} \times \real \times \natural^{+}$
  to denote the domain of observations.
  An entire $D$-dimensional time series can thus be represented as
  \begin{equation}
    \mathcal{S}_i := \left\{\left(t_{1}, z_{1}, m_{1}\right), \dots, \left(t_{M}, z_{M}, m_{M}\right)\right\},
  \end{equation}
  where for notational convenience we omitted the index $i$ from
  individual measurements.
  \label{def:Time series}
\end{definition}
We leave this definition very general on purpose, in particular allowing
the length of each time series to \emph{differ}, since our models are
inherently capable of handling this.
Likewise, we neither enforce
nor expect all time series to be synchronized, i.e.\ being sampled at
the same time, but rather we are fully agnostic to \emph{non-synchronized}
observations in the sense of not having to observe all modalities at
each time point\footnote{We make no assumptions about the time values
$t_j$ and merely require them to be positive real-valued numbers because
our time encoding procedure~(see below) is symmetric with respect to
zero. In practice, positive time values can always be achieved by
applying a shift transformation.}.
We collect all time series and their associated labels in a dataset
$\mathcal{D}$.
\begin{definition}[Dataset]
  We consider a dataset $\mathcal{D}$ to consist of $n$ time series.
  Elements of $\mathcal{D}$ are tuples, i.e.\
  $ \mathcal{D} := \mleft\{\mleft(\mathcal{S}_1, y_1\mright), \dots, \mleft(\mathcal{S}_N,
  y_N\mright)\mright\}$, where $\mathcal{S}_i$ denotes the $i$\th time
  series and $y_i \in \{1, \dots, C\}$ its class label.
  \label{def:Dataset}
\end{definition}
For an online monitoring scenario, we will slightly modify
Definition~\ref{def:Dataset} and only consider \emph{subsets} of time
series that have already been observed.
\prettyref{fig:Workflow} gives a high-level overview of our method,
including the individual steps required to perform classification.
To get a more intuitive grasp of these definitions, we briefly
illustrate our time
series notation with an example.
Let instance $i$ be an in-hospital patient, while the time series
represent measurements of two channels of vital parameters during
a hospital stay, namely heart rate~(HR) and mean arterial blood
pressure~(MAP). We enumerate those channels as modalities~1 and~2.
Counting from admission time, a HR of 60 and 65 beats per minute~(bpm) was
measured after \SI{0.5}{\hour} and \SI{3.0}{\hour}, respectively,
whereas MAP values of 80, 85, and \SI{87}{\mmHg} were observed after
\SI{0.5}{\hour}, \SI{1.7}{\hour}, and \SI{2.5}{\hour}.
According to Definition~\ref{def:Time series}, the time series is thus
represented as
$
  \mathcal{S}_i := \{
    (0.5, 60, 1 ),
    (3, 65, 1 ),\allowbreak
    (0.5, 80, 2 ),\allowbreak
    (1.7, 85, 2 ),\allowbreak
    (3, 87, 2 )
  \}.
$
In this example, observations are ordered by modality to increase
readability; in practice, we are dealing with unordered sets.
This does \emph{not} imply, however, that we ``throw away'' any time
information; we encode time values in our model, thus making it possible
to maintain the temporal ordering of observations. Our model, however,
does not assume that all observations are stored or processed in the
same ordering---this assumption was already
shown~\citep{vinyals2015order} to be detrimental with respect to
classification performance in some scenarios.
Therefore, our model does not employ a ``sequentialness prior'':
instead of processing a sequence conditional on previously-seen
elements~(such as in RNNs or other sequence-based models), it
processes values of a sequence \emph{all at once}---through encoding and
aggregation steps---and retains all information about event occurrence
times.

In our experiments, we will focus on time series in which certain
modalities---channels---are not always observed, i.e.\ some
measurements might be missing. We call such time series
\emph{non-synchronized}.
\begin{definition}[Non-synchronized time series]
  A \mbox{$D$-dimensional} time series is \emph{non-synchronized} if there
  exists at least one time point $t_{j} \in \real^{+}$ at which at least one
  modality is not observed, i.e.\ if there is $t_{j} \in \real^{+}$
  such that $\left|\left\{\left(t_{k}, z_{k}, m_{k} \right) \mid
  t_{k}=t_{j} \right\}\right| \neq D$.
\end{definition}
Furthermore, we require that the measurements of each modality satisfy
$t_i \neq t_j$ for $i \neq j$, such that no two measurements of of the
same modality occur at the same time.  This assumption is not required
for technical reasons but for consistency; moreover, it permits
\emph{interpreting} the results later on. 

\paragraph{Summary} To summarize our generic setup, we do not require
$M$, the number of observations per time series, to be the same, i.e.\
$|\mathcal{S}_i| \neq |\mathcal{S}_{j}|$ for $i \neq j$ is
permitted, nor do we assume that the time points and modalities of the
observations are the same across time series.
This setting is common in clinical and biomedical time series. Since
typical machine learning algorithms are designed to operate on data of
a \emph{fixed} dimension, novel approaches to this non-trivial problem
are required.

\subsection{Our Model}

In the following, we describe an approach inspired by differentiable learning of
functions that operate on sets~\citep{zaheer2017deep, Wagstaff2019}. The
following paragraphs provide a brief overview of this domain, while
describing the building blocks of our model.
Specifically, we  phrase the problem of classifying time series on irregular grids as learning
a function $f$ on a set of arbitrarily many time series observations following
Definition~\ref{def:Time series}, i.e.\
$\mathcal{S} = \{(t_{1}, z_{1}, m_{1}), \dots, (t_{M}, z_{M}, m_{M})\}$,
such that
$f\colon \mathcal{S} \to \real^C$,
where $\mathcal{S}$ represents a generic time series of arbitrary
cardinality and $\mathbb{R}^C$ corresponds to the logits of the $C$
classes in the dataset.
As we previously discussed, we interpret each time series as an unordered set of
measurements, where all information is conserved because the observation
time is included for each set element. Specifically, we define $f$ to be
a set function, i.e.\ a function that operates on a set and thus has to be
\emph{invariant} to the ordering of the elements in the set.
Multiple architectures are applicable to constructing set functions such as
Transformers~\citep{lee2019set, vaswani2017attention}, or Deep
Sets~\citep{zaheer2017deep}. Given its exceptional scalability
properties, we base this work on the framework of
\citet{zaheer2017deep}. Intuitively, this amounts to computing
multivariate dataset-specific summary statistics, which are optimized to
maximize classification performance. Thus, we \emph{sum-decompose} the
set function~$f$ into the form
\begin{equation}
    f(\mathcal{S})
        = g\mleft(\frac{1}{|\mathcal{S}|} \sum_{s_j \in \mathcal{S}} h\mleft(s_j\mright) \mright)
    \label{eq:sum-decomposed-set-fn}
\end{equation}
where $h\colon\Omega\to\real^d$ and $g\colon\real^d\to\real^C$ are
neural networks, $d \in \natural^{+}$ determines the
dimensionality of the latent representation, and $s_j$
represents a single observation of the time series $\mathcal{S}$.
We can view the averaged representations $1 / |\mathcal{S}| \sum_{s_j
\in \mathcal{S}} h(s_j)$ in general as a dataset-specific summary
statistic learned to \emph{best} distinguish the class labels.
\prettyref{eq:sum-decomposed-set-fn} also implies the beneficial
scalability properties of our approach: each embedding can be
calculated independently of the others; hence, the constant
computational cost of passing a single observation through the
function $h$ is scaled by the number of observations, resulting in
a runtime of $\mathcal{O}(M)$ for a time series of length $M$.

Recently, \citet{Wagstaff2019} derived requirements for a practical universal
function representation of \emph{sum-decomposable} set functions, i.e\ the
requirements \emph{necessary} for a \emph{sum-decomposable} function to
represent an arbitrary set-function given that $h$ and $g$ are arbitrarily
expressive.  In particular, they show that a universal function representation
can only be guaranteed provided that
$d \geq \max_i |\mathcal{S}_i|$
is satisfied.  
During hyperparameter search, we therefore independently sample the
dimensionality of the aggregation space, and allow it to be in the order
of the number of observations that are to be expected in the dataset.
Further, we explored the utilization of $\text{max}$, $\text{sum}$, and
$\text{mean}$ as alternative aggregation functions inspired by
\citet{zaheer2017deep,garnelo2018conditional}.

Intuitively, our method can be related to
Takens's embedding theorem~\citep{Takens81} for dynamical systems:
we also observe a set of samples from some unknown~(but deterministic)
dynamical process; provided the dimensionality of our architecture is
sufficiently large\footnote{%
  In Takens's embedding theorem, $d > d_B$ is required, where $d_B$
  refers to the fractal box counting dimension~\citep{Liebovitch89},
  which is typically well below the size of typical neural network
  architectures.
}, we are capable of reconstructing the system up to diffeomorphism.
The crucial difference is that we do \emph{not} have to construct
a time-delay embedding~(since we are not interested in being able to
perfectly reproduce the dynamics of the system) but rather, we let the
network learn an embedding that is suitable for classification, which is
arguably a simpler task due to its restricted scope.

\paragraph{Time Encoding}
To represent the time point of an observation on a normalized scale,
we employ a variant of \emph{positional encodings}~\citep{vaswani2017attention}.
Preliminary results indicated that this encoding scheme reduces the
sensitivity towards initialization and training hyperparameters of
a model.
Specifically, the time encoding converts the \mbox{$1$-dimensional} time
axis into a multi-dimensional input by passing the time $t$ of each
observation through multiple trigonometric functions of varying
frequencies.
Given a dimensionality $\tau \in \natural^{+}$ of the time encoding, we
refer to the encoded position as $x \in \real^{\tau}$, where
\begin{align}
  x_{2k}(t)   &:= \sin\left( \frac{t}{\mathfrak{t}^{2k / \tau}} \right)\\
  x_{2k+1}(t) &:= \cos\left( \frac{t}{\mathfrak{t}^{2k / \tau}} \right)
\end{align}
with $k \in \{0, \dots, \nicefrac{\tau}{2} \}$ and $\mathfrak{t}$
representing the maximum time scale that is expected in the
data.
Intuitively, we select the wavelengths using a geometric
progression from $2\pi$ to $\mathfrak{t} \cdot 2\pi$, and treat the
number of steps and the maximum timescale $\mathfrak{t}$ as
hyperparameters of the model.  We used time encodings for all
experiments, such that an observation is represented as $s_j
= \left(x\left(t_j\right), z_j, m_j \right)$.

\subsection{Attention-based Aggregation}\label{sec:attention}

So far, our method permits encoding sets of arbitrary sizes into
a fixed-size representation. For increasingly large set sizes, however,
many irrelevant observations could influence the result of the set
function. The \emph{mean} aggregation function is particularly
susceptible to this because the influence of an observation to the
embedding shrinks proportionally to the size of the set.
We thus suggest to use a \emph{weighted} mean in order to allow the
model to decide which observations are relevant and which should be
considered irrelevant. This is equivalent to computing an attention
over the set input elements, and
subsequently, computing the sum over all elements in the set.

Our approach is based on \emph{scaled dot-product attention} with
multiple heads $i \in \{1, \dots, m\}$ in order to be able to cover
different aspects of the aggregated set\footnote{%
  Since we are dealing only with a single instance~(i.e.\ time series)
  in this section, we use $i$ and $j$ to denote
  a \emph{head} and an \emph{observation}, respectively.
}.
We define $\attention(\mathcal{S}, s_j)$, i.e.\ the attention weight
function of an individual time series, to depend on the overall set
of observations~$\mathcal{S}$, and the value of the set element~$s_j$.
This is achieved by computing an embedding of the set
elements using a smaller set function $f'$, and projecting the
concatenation of the set representation and the individual set elements
into a \mbox{$d$-dimensional} space. Specifically, we have
%
    $K_{j, i} = [f'(\mathcal{S}), s_j]^T W_i$
%
where $W_i \in \real^{(\im(f') + |s_j|)\times d}$ and $K \in \real^{|\mathcal{S}| \times d}$.
Furthermore, we define a matrix of query points $Q \in \real^{m \times
d}$, which allow the model to summarize different aspects of the dataset
via
\begin{equation*}
  e_{j, i} = \frac{K_{j,i} \cdot Q_{i}}{\sqrt{d}} \quad\quad\text{and}\quad\quad \attention_{j, i} = \frac{\exp(e_{j,i})}{\sum_j \exp(e_{j,i})}
\end{equation*}
where $\attention_{j, i}$ represents the amount of attention that head $i$
gives to set element $j$.
The head-specific row $Q_i$ of the query matrix $Q$ allows a head to
focus on individual aspects~(such as the distribution of one or multiple
modalities) of a time series.
For each head, we multiply the set element embeddings
computed via the function $h$ with the attentions derived for the
individual instances, i.e.\ $r_{i} = \sum_j \attention_{j,i} h(s_j)$.
The computed representation is concatenated and passed to the aggregation
network $g_\theta$ as in a regular set function, i.e.\
$r* = [r_1 \dots r_m]$.
In our setup, we initialize $Q$ with zeros, such that at the beginning of
training, the attention mechanism is equivalent to computing the unweighted
mean over the set elements.

Overall, this aggregation function is similar to Transformers
\citep{vaswani2017attention}, but differs from them in a few key
aspects. Commonly, Transformer blocks use the information from
\emph{all set elements} to compute the embedding of an individual set element,
leading to a runtime and space complexity of~$\mathcal{O}(n^2)$. By contrast,
our approach computes the embeddings of set elements independently, leading
to lower runtime and memory complexity of $\mathcal{O}(n)$. This is
particularly relevant as set elements in our case are \emph{individual
observations}, so that we obtain  set sizes that are often multiples of the
time series length. Furthermore, we observed that computing embeddings
with information from other set elements~(as the Transformer does)
actually \emph{decreases} generalization performance in several
scenarios~(please refer to \prettyref{tab:results} for details).

\paragraph{Online monitoring scenario}
In an online monitoring scenario, we compute all variables of the model in
a cumulative fashion.  The set of observations used to predict at the current
time point is therefore a subset of the total observations available at the time
point at which the prediction is made.  If this were computed na\"ively, the
attention computation would result in $\mathcal{O}(|\mathcal{S}|)$ runtime and
memory complexity, where $|\mathcal{S}|$ is the number of observations.
Instead we \emph{rearrange} the computation of the weighted mean as
follows, while discarding the head indices for simplicity:
\begin{align*}
    f(\mathcal{S}_i) &= \sum_{j \leq i} \frac{\exp(e_j)}{\sum_{k \leq i} \exp(e_k)} h(s_j) \\
        &= \frac{\sum_{j \leq i} \exp(e_j) h(s_j) }{\sum_{k \leq i} \exp(e_k)}
\end{align*}
In the previous equation, both numerator and denominator can be computed
in a cumulative fashion and thus allow reusing computations from
previous time points.

\subsubsection{Loss Function}
%
If not mentioned otherwise, we choose~$h$ and~$g$ in
\prettyref{eq:sum-decomposed-set-fn} to be \emph{multilayer perceptron}
deep neural networks, parametrized by weights $\theta$ and $\psi$,
respectively. We thus denote these neural networks by $h_\theta$ and
$g_\psi$; their parameters are shared across all instances per dataset. 
Our training setup follows \citet{zaheer2017deep}; we apply the set
function to the complete time series, i.e.\ to the set of all
observations for each time series.
Overall, we optimize a loss function that is defined as
\begin{equation*}
  \begin{medsize}
    \mathcal{L}(\theta, \psi) := \E_{(\mathcal{S}, y) \in \mathcal{D}} \left[
        \ell \left(
            y; g_\psi \left(
              \sum_{s_j \in \mathcal{S}} \attention(\mathcal{S}, s_j) h_\theta(s_j)
            \right)
        \right)
    \right],
  \end{medsize}
\end{equation*}
where $\ell(\cdot)$ represents a task-specific loss function. In all of our
experiments, we utilize the binary cross-entropy loss in combination with
a sigmoid activation function in the last layer of $g_\psi$ for binary
classification.

\section{Experiments}

We executed all experiments and implementations in a unified and modular code
base, which we make available to the community. We provide two dedicated
packages
\begin{inparaenum}[(i)]
    \item for automatic downloading and preprocessing of the datasets according to the splits
used in this work and
    \item for training the introduced method and baselines to which we compare
        in the following.
\end{inparaenum}
We make both publicly available\footnote{\url{https://github.com/BorgwardtLab/Set_Functions_for_Time_Series}}.
While some of the datasets used in the following have access restrictions,
anybody can gain access after satisfying the defined requirements.
This ensures the reproducibility of our results. Please consult
\prettyref{appx:imp-details} for further details.

\subsection{Datasets}

In order to benchmark the proposed method, we selected three datasets
with irregularly-sampled and non-synchronized measurements. We are
focusing on two tasks with different challenges: first, we predict
patient mortality on two datasets; this task is exacerbated by the high
imbalance in the datasets. Second, we predict the onset of
\emph{sepsis}\footnote{An organ dysfunction caused by a dysregulated
host response to an infection. Sepsis is potentially life-threatening
and is associated with high mortality of patients.} in an online
scenario.

\paragraph{MIMIC-III Mortality Prediction}
%
MIMIC-III \citep{johnson2016mimic} is a widely-used, freely-accessible
dataset consisting of distinct ICU stays of patients. The median length
of a stay is \SI{2.1}{\day}; a wide range of physiological
measurements~(e.g.\ MAP and HR) are recorded with a resolution of
\SI{1}{\hour}.
Furthermore, laboratory test results, collected at irregular time
intervals, are available.
Recently, \citet{Harutyunyan2019} defined a set of machine learning tasks,
labels, and benchmarks using a subset of the MIMIC-III dataset.
We trained and evaluated our method and competing methods on the binary
mortality prediction task~(\dataset{M3-Mortality}), while discarding the
binning step and applying additional filtering
described in \prettyref{appx:data-filtering}.
The goal of the mortality prediction task~(which we abbreviate as
\dataset{M3-Mortality}) is to predict whether
a patient will die during their hospital stay using only data from the
first \SI{48}{\hour} of the ICU stay. In total, the dataset contains
around $21,000$ stays of which approximately \SI{10}{\percent} result in
death.

\paragraph{Physionet 2012 Mortality Prediction Challenge}
%
The 2012 Physionet challenge dataset~\citep{goldberger2000physiobank},
which we abbreviate \dataset{P-Mortality}, contains $12,000$ ICU stays
each of which lasts at least \SI{48}{\hour}.
For each stay, a set of general descriptors~(such as gender or age)
are collected at admission time.
Depending on the course of the stay and patient status, up to $37$ time series
variables were measured~(e.g.\ blood pressure, lactate, and respiration
rate).
While some modalities might be measured in regular time intervals~(e.g.
hourly or daily), some are only collected when required; moreover, not
all variables are available for each stay. 
Similar to \dataset{M3-Mortality}, our goal is to predict whether a patient
will die during the hospital stay. 
The training set comprises $8,000$ stays, while the testing set
comprises $4,000$ ICU visits.
The datasets are similarly imbalanced, with a prevalence of around
\SI{14}{\percent}.
Please refer to \prettyref{tab:physionet prevalence}, \prettyref{tab:mimic
binary prevalence}, and \prettyref{tab:sepsis prevalence} in the
appendix for a more detailed enumeration of samples sizes, label
distributions, and the handling of demographics data.
The total number of samples slightly deviates from the
originally-published splits, as time series of excessive length
precluded us from fitting some methods in reasonable time, and we therefore
excluded them.

\paragraph{Physionet 2019 Sepsis Early Prediction Challenge}
%
Given the high global epidemiological burden of sepsis,
\citet{reyna2020early} launched a challenge for the early detection of
sepsis from clinical data. Observations from over $60,000$ intensive
care unit patients from three different U.S.\ hospitals
were aggregated. Up to $40$ variables~(e.g.\ vitals signs and lab results)
were recorded hourly, with each hour being labelled with a binary
variable indicating whether an onset of sepsis---according to the
\mbox{Sepsis-3} definition~\citep{seymour2016assessment}---occurred.
Our task is the hourly prediction of sepsis onset within the next
\SIrange{6}{12}{\hour}. In our work we refer to this task as \dataset{P-Sepsis}.
To account for clinical utility of a model, the authors introduced a novel evaluation 
metric~(see \citet{reyna2020early} for more details), which we also
report in our experiments: at each prediction time point $t$, a classifier 
is either \emph{rewarded} or \emph{penalized} using a utility function
$U(p,t)$, which depends on the 
distance to the actual sepsis onset for patient~$p$. The total utility function is the sum over 
all patients $P$ and the predictions at all time points $T$, i.e.\
$U_{\text{total}} := \sum_{p \in P}\sum_{t \in T}U(p,t)$.
The score is then normalized ($U_{\text{norm}}$) such that a perfect classifier receives
a score of~$1$, while a classifier with \emph{no} positive predictions
at all receives a score of~$0$.

\subsection{Competitor Methods}

In order to achieve a thorough comparison, we compare our method to the
following six approaches: \begin{inparaenum}
    \item GRU-simple~\citep{che2018recurrent}
    \item GRU-Decay ~\citep{che2018recurrent}
    \item Phased-LSTM~\citep{neil2016phased}
    \item Interpolation Prediction Networks~\citep{shukla2018interpolationprediction}
    \item Transformer~\citep{vaswani2017attention}
    \item Latent-ODE~\citep{rubanova2019latent}
\end{inparaenum}.

All methods except \method{Latent-ODE} were implemented in the same framework
using the same training pipeline.
Due to differences in implementation and limited comparability, we highlight
the results of \method{Latent-ODE} with $^\dagger$.  In particular, for
LatentODE we were unable to run an extensive hyperparameter search using the
provided codebase, as runtime was considerable higher compared to any
other method.  This results in an \emph{underestimation} of performance for
\method{Latent-ODE} compared to the other methods. For a detailed description
of the differences between the implementations, we refer the reader to
\prettyref{appx:imp-details}.

\subsection{Experimental Setup}

To mitigate the problem of unbalanced datasets, all models were trained on
balanced batches of the training data rather than utilizing class weights. This
was done in order to not penalize models with a higher memory footprint\footnote{These
models would only allow training with small batch sizes, which combined with
the unbalanced nature of the datasets would lead to high variance in the
gradients.}. Due to oversampling, the notion of an epoch is different from
common understanding. In our experiments we set the number of optimizer steps
per epoch to be the minimum of the number of steps required for seeing all
samples from the majority class and the number of steps required to see each
samples from the minority class three times.
Training was stopped after $30$ epochs without improvement of the area under
the precision--recall curve~(AUPRC) on the validation data for the mortality
prediction tasks, whereas balanced accuracy was utilized for the online
predictions scenario. The hyperparameters with the best overall validation
performance were selected for quantifying the performance on the test set. The
train, validation, and test splits were the same for all models and all
evaluations.
To permit a fair comparison between the methods, we executed hyperparameter
searches for each model on each dataset, composed of uniformly sampling $20$
parameters according to \prettyref{appx:hyperparameter-search}.
Final performance on the test set was calculated by~$3$ \emph{independent} runs
of the models; evaluation took place after the model was restored to the state
with the best validation AUPRC / balanced accuracy. In all subsequent
benchmarks, we use the standard deviation of the test performance of these runs
as generalization performance estimates.

\subsection{Results}\label{sec:results}

\begin{figure}[t]
  \centering
  \begin{tikzpicture}[scale=0.5]
    \begin{groupplot}[%
      group style = {%
        group size     = 2 by 1,
        ylabels at     = edge left,
        xlabels at     = edge bottom,
        xticklabels at = edge bottom,
        vertical sep   = 0.50cm,
      },
      nodes near coords,
      axis x line*  = bottom,
      axis y line*  = left,
      xlabel        = {AUPRC},
      ylabel        = {Runtime in \si{\second}},
      enlargelimits = true,
    ]
      \nextgroupplot[title={\dataset{M3-Mortality}}]
        \addplot[only marks, mark=*, point meta = explicit symbolic] coordinates {
            (52.04,  133.29) [\tiny\method{GRU-D}]
            (43.60,  139.74) [\tiny\method{GRU-Simple}]
            (48.31,   81.17) [\tiny\method{IP-Nets}]
            (37.05,  165.50) [\tiny\method{Phased-LSTM}]
            (42.55,   20.11) [\tiny\method{Transformer}]
            (46.28,  14.51) [\tiny\bfseries\method{\methodname-Attn}]
        };

      \nextgroupplot[title={\dataset{P-Mortality}}]
        \addplot[only marks, mark=*, point meta = explicit symbolic] coordinates {
            (53.74,   8.57) [\tiny\method{GRU-D}]
            (42.18,  38.95) [\tiny\method{GRU-Simple}]
            (51.04,  25.25) [\tiny\method{IP-Nets}]
            (52.81,   6.06) [\tiny\method{Transformer}]
            (38.73,  44.60) [\tiny\method{Phased-LSTM}]
            (52.40,   7.62) [\tiny\bfseries\method{\methodname-Attn}]
        };
    \end{groupplot}
  \end{tikzpicture}
  \caption{
    Runtime vs.\ AUPRC trade-offs for all methods on the two mortality
    prediction tasks.  \method{Latent-ODE} is not shown as its runtime is
    significantly higher compared to the other models.
  }
  \label{fig:Runtime}
\end{figure}
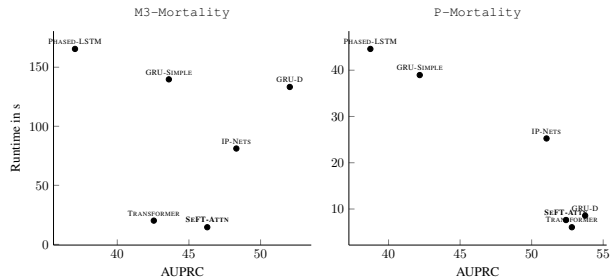

\prettyref{tab:results} depicts the results on the two mortality
prediction tasks. For each metric, we use bold font to indicate the best
value, and italics to indicate the second-best. Overall, our proposed
method \method{\methodname-Attn} exhibits competitive performance. In
terms of AUPRC, arguably the most appropriate metric for unbalanced
datasets such as these, we consistently rank among the first four
methods.
For \dataset{M3-Mortality}, abbreviated as \dataset{M3M} in the table, our
runtime performance is lower than that of \method{Transformer}, but we
outperform it in terms of AUPRC.  Here both \method{GRU-D} and \method{IP-Nets}
outperform the devised approach, while exhibiting considerably higher runtimes.
The favourable trade-offs of \method{\methodname-Attn} between runtime and
AUPRC are further underscored by \prettyref{fig:Runtime}.
On \dataset{P-Mortality}, abbreviated as \dataset{P12} in the table, our
method is comparable to the performance of \method{GRU-D} and
\method{Transformer} and shows comparable or lower runtime.  This
picture is similar for the Area under the ROC curve~(AUROC), where
\method{IP-Nets} show a slightly higher performance than our approach, at
a cost of almost three-fold higher runtime.

\begin{table}[tbp]
\caption{%
    Performance comparison of methods on mortality prediction datasets.
    ``\textsc{AUROC}'' denotes the area under the Receiver Operating
    Characteristic~(ROC) curve;
    ``\textsc{AUPRC}'' denotes the area under the precision--recall curve.
    Evaluation metrics were scaled to $100$ in order to increase readability.
    $^\dagger$ denotes that the performance could be underestimated due to
    limited hyperparameter tuning compared to other methods.
}
\vspace{0.10in}
\label{tab:results}
\centering
\input{tables/the_big_boss}
\end{table}

\paragraph{Opening the black box}

\begin{figure*}[tb]
    \centering
    \includegraphics[width=\linewidth]{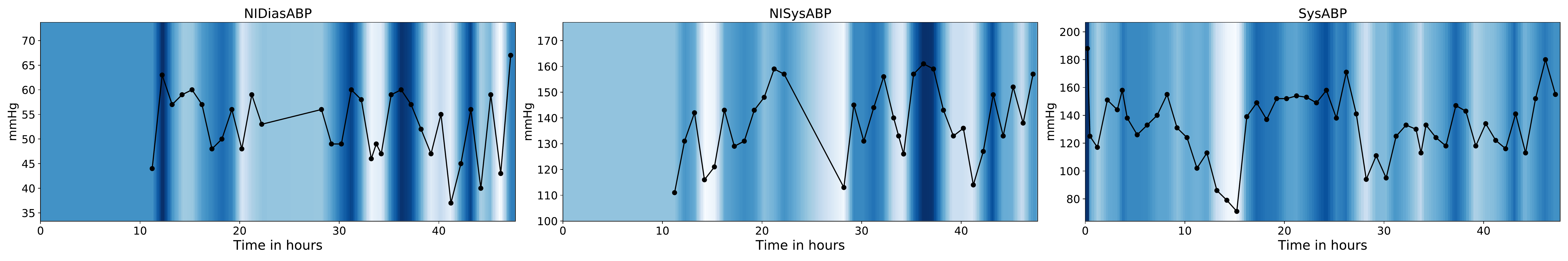}
    \caption{%
      Visualizations of a single attention head on an instance of the
      \dataset{P-Mortality} dataset. We display a set of blood pressure
      variables which are most relevant for assessing patient stability: 
      non-invasive diastolic arterial blood pressure (NIDiasABP),
      non-invasive systolic arterial blood pressure (NISysABP),
      and invasively measured systolic arterial blood pressure~(SysABP).
      Darker colors represent higher attention values. In the invasive channel
      showing high time resolution (right most panel), our model attends
      foremost to the region after a sudden increase in blood pressure. In the
      non-invasive, intermittently observed channels, the model additionally
      focuses on regions of high observation density reflecting the clinicians
      concern.
    }
    \label{fig:attention}
\end{figure*}

In the medical domain, it is of particular interest to understand the decisions
a model makes based on the input it is provided with. The formulation of our
model and its \emph{per-observation} perspective on time series gives it the
unique property of being able to quantify to which extent an individual
observation contributed to the output of the model.
We exemplify this in \prettyref{fig:attention} with a patient time
series that was combined with the attention values of our model, displayed for a set
of clinically relevant variables.
After reviewing these records with our medical expert, we find that in channels
showing frequent and regularly-spaced observations, the model attends to
drastic changes. For instance, see the sudden increase in continuously-monitored
invasive systolic blood pressure. Interestingly, in channels that are
observed only intermittently~(due to manual intervention, such as non-invasive
blood pressure measurements), we observe that our model additionally attends to
regions of high observation density, thereby reflecting the increased
concern of clinicians for the circulatory stability of patients.

\paragraph{Online Prediction Scenario}

In order to provide models with potential clinical applicability, it is
instrumental to cover online monitoring scenario to potentially support
clinicians. We present the results of the Sepsis early prediction online
monitoring scenario \dataset{P-Sepsis} in \prettyref{tab:online}.
In this scenario the \method{Transformer} and \method{IP-Nets} yield the
highest performance and outperform all other methods almost two-fold.  These
results are very surprising, given that the best out of 853 submissions to the
Physionet 2019 challenge only achieved a test utility score of
$0.36$~\citep{reyna2020early}.  In order to investigate this issue further, we
designed a second evaluation scenario, where future information of each
instance is guaranteed to be unavailable to the model, by splitting each
instance into multiple cumulative versions of increasing length.  We then ran
all trained models in this scenario and included results for models where the
performance metrics differ in \prettyref{tab:online}, highlighted with an
additional~$^*$.
It is clearly recognizable that the performance of both \method{IP-Nets}
and \method{Transformer} decrease in the second evaluation scenario
indicating the models' reliance on leaked future information. 

For \method{IP-Nets}, information can leak through the non-parametric imputation
step prior to the application of the downstream recurrent neural network.  It
is infeasible to train the vanilla \method{IP-Nets} approach on slices of the
time series up until the time point of prediction, as we cannot reuse
computations from previous imputation steps.  While it would be possible to
construct an \method{IP-Nets} variant that does \emph{not} rely on future information
during the imputation step, for example using smoothing techniques, we deem
this beyond the scope of this work.

Similar effects occur in the case of the \method{Transformer}: While
observations from the future are masked in the attention computation,
preventing access to future values results in a detrimental reduction in
performance. Even though the source of dependence on future values is quite
probable to reside in the layer normalization applied in the
\method{Transformer} model, the performance drop can have multiple
explanations, i.e.\
\begin{inparaenum}
\item the absence of future time points leads to high variance estimates of
    mean and variance in the layer norm operation, resulting in bad performance
    in the initial time points of the time series, or
\item the model actively exploits future information though the layer
    normalization.  This could for example be possible by the model looking for
    indicative signals in future time points and when present returning very
    high norm outputs.  The signature of these high norm outputs can then,
    through the layer norm operation, be observed in earlier time points.
\end{inparaenum}
While one could construct variants where the \method{Transformer} model can by
no means access future information, for example by replacing the layer norm
layer with and alternative normalization scheme~\citep{nguyen2019transformers,
bachlechner2020rezero}, we reserve a more thorough investigation of this issue
for future work.

By contrast, our model does not contain any means of leaking future
information into the prediction of the current time point and thus
exhibits the same performance in both evaluation scenarios,  while
remaining competitive with alternative approaches.
Surprisingly, the model with the highest performance in this scenario is
\method{GRU-Simple}, which could be explained by the very regular sampling
character of the \dataset{P-Sepsis} dataset.  Here the measurements were
already binned into hours, such that it cannot be considered completely
irregularly sampled.  This explains the high performance of
\method{GRU-Simple}, as compared to models which were specifically
designed to cope with the irregular sampling problem.

\begin{table}[]
    \centering
    \caption{%
    Results of the online prediction scenario on the \dataset{P-Sepsis}
    task. The dataset is highly imbalanced, such that we only report
    measures which are sensitive to class imbalance.  Further, if the results
    between the evaluation scenarios differ, we highlight results
    \emph{without} masked future information in \textcolor{gray}{gray}, and the
    performance achieved with masking with $^*$. $^\dagger$ indicates that the
    results might be underestimating the true performance due to limited
    hyperparameter tuning compared to the other methods.
    } %
    \vspace{0.10in}
    \input{tables/online.tex}
    \label{tab:online}
\end{table}

\section{Conclusion and Discussion}

We presented a novel approach for classifying time series with
irregularly-sampled and unaligned---i.e.\ \emph{non-synchronized}---observations.
Our approach yields competitive performance on real-world datasets with low
runtime compared to many competitors.  While it does not out perform
state-of-the-art models, it shows that shifting the perspective to individual
observations represents a promising avenue for models on irregularly-sampled
data in the future.  Further, as the model does not contain any
``sequential'' inductive biases compared to RNNs, it indicates that for time series
classification tasks, this bias might not be of high importance. This is in
line with recent research on Transformer architectures for NLP
\cite{vaswani2017attention}, where order is solely retained through positional
embeddings and not inherent to the processing structure of the model.
Our experiments demonstrated that combining the perspective of individual
observations with an attention mechanism permits increasing the
interpretability of the model. This is particularly relevant for
medical and healthcare applications.

Along these lines, we also want to briefly discuss a phenomenon that we
observed on \dataset{M3-Mortality}: the performance we report on this task is
often \emph{lower} than the one reported in the original
paper~\citep{Harutyunyan2019} and follow-up work~\citep{song2018attend}.
We suspect
that this is most likely caused by a distribution shift between the validation
dataset and the test dataset: in fact, model performance is on average 6.3\%
higher~(in terms of AUPRC) on the validation data set than on the test dataset,
which might indicate that the validation dataset is not representative here.
We also notice that previous work~\citep{Harutyunyan2019} uses
heavily-regularised and comparatively small models for this specific
scenario, making them more impervious to distribution shifts.  Finally, the
fact that we do not bin the observations prior to the application of the models
could make our task more difficult compared to the original setup.

Additionally, we would like to discuss the low performance of the
\method{Latent-ODE} model. This can be partially attributed to the fact that
we did not perform an extensive hyperparameter search with this model.
Furthermore, all runs of the \method{Latent-ODE} model contain an additional
reconstruction loss term, so that it is necessary to define the trade-off
between reconstruction and classification.  In this work we used the same
parameters as \citet{rubanova2019latent}, which~(due to potentially different
characteristics of the datasets) could lead to a less than optimal trade-off.
For example, some modalities in the datasets we considered might be harder to
reconstruct, such that the value of the reconstruction term could be higher,
leading to gradients which prioritize reconstruction over classification.
Nevertheless, despite these differences the performance of \method{Latent-ODE}
in terms of AUC on the \dataset{P-Mortality} dataset measured in this work is
actually \emph{higher} than the performance reported in the original publication.

In future work, we plan to explore attention mechanisms specifically designed
for sets of very high cardinalities. We also strive to make the attention
computation more robust so that elements with low attentions values do not get
neglected due to numerical imprecision of aggregation operations; this is also
known as \emph{catastrophic cancellation}~\citep{Knuth98}, in our case,
\emph{summation}.
GPU implementations of algorithms such as Kahan summation~\citep{Kahan65} would
represent a promising avenue for further improving performance of attention
mechanisms for set functions.

\subsection*{Acknowledgements}
The authors would like to thank Dr.\ Felipe Llinares López for comments and
discussions during the progress of this work.

This project was supported by the grant \#2017‐110 of the Strategic
Focal Area ``Personalized Health and Related Technologies (PHRT)'' of
the ETH Domain for the SPHN/PHRT Driver Project ``Personalized Swiss
Sepsis Study'' and  the SNSF Starting Grant ``Significant Pattern
Mining''~(K.B., grant no.~155913).
Moreover, this work was funded in part by the Alfried Krupp Prize for Young
University Teachers of the Alfried Krupp von Bohlen und
Halbach-Stiftung~(K.B.).

\bibliography{references}
\bibliographystyle{icml2020}

\counterwithin{figure}{section}
\counterwithin{table} {section}

\clearpage
\appendix

\section{Appendix}

\begin{table}[h]
    \centering
    \caption{\dataset{M3-Mortality} prevalence of labels for the binary classification task}
    \vspace{0.10in}
    \input{tables/mimic_binary_prevalences}
    \label{tab:mimic binary prevalence}
\end{table}

\begin{table}[h]
    \centering
    \caption{\dataset{P-Mortality} prevalence of labels for the binary classification task}
    \vspace{0.10in}
    \input{tables/physionet_binary_prevalences}
    \label{tab:physionet prevalence}
\end{table}

\begin{table}[ht]
    \centering
    \caption{\dataset{P-Sepsis} prevalence of labels for the online
      prediction task}
    \vspace{0.10in}
    \begin{tabularx}{\linewidth}{llll}
     \toprule
      & Train & Test & Val\\ \midrule
      Sepsis occurrence & \num{0.018} & \num{0.018} & \num{0.018}\\
    \bottomrule
    \end{tabularx}
    \label{tab:sepsis prevalence}
\end{table}

\subsection{Dataset preprocessing} \label{appx:data-filtering}
\paragraph{Filtering}
Due to memory requirements of some of the competitor methods, it was necessary
to excluded time series with an extremely large number of measurements. 
For \dataset{M3-Mortality}, patients with more than 1000 time points 
were discarded as they contained dramatically different measuring frequencies 
compared to the rest of the dataset. This led to the exclusion of the
following 32 patient records:
\texttt{73129\_2}, \texttt{48123\_2}, \texttt{76151\_2}, \texttt{41493\_1},
\texttt{65565\_1}, \texttt{55205\_1}, \texttt{41861\_1}, \texttt{58242\_4},
\texttt{54073\_1}, \texttt{46156\_1}, \texttt{55639\_1}, \texttt{89840\_1},
\texttt{43459\_1}, \texttt{10694\_2}, \texttt{51078\_2}, \texttt{90776\_1},
\texttt{89223\_1}, \texttt{12831\_2}, \texttt{80536\_1}, \texttt{78515\_1},
\texttt{62239\_2}, \texttt{58723\_1}, \texttt{40187\_1}, \texttt{79337\_1},
\texttt{51177\_1}, \texttt{70698\_1}, \texttt{48935\_1}, \texttt{54353\_2},
\texttt{19223\_2}, \texttt{58854\_1}, \texttt{80345\_1}, \texttt{48380\_1}.

In the case of the \dataset{P-Mortality} dataset, some instances did not
contain any time series information at all and were thus removed. This led to
the exclusion of the following 12 patients:
\texttt{140501}, \texttt{150649}, \texttt{140936}, \texttt{143656},
\texttt{141264}, \texttt{145611}, \texttt{142998}, \texttt{147514},
\texttt{142731},\texttt{150309}, \texttt{155655}, \texttt{156254}.

For \dataset{P-Sepsis} some instances did not contain static values or were
lacking time series information all together. We thus excluded the following
files: \texttt{p013777.psv}, \texttt{p108796.psv}, \texttt{p115810.psv}.

\paragraph{Static variables}

The datasets often also contain information about static variables, such
as age and gender. \prettyref{tab:static variables} lists all the static
variables for each of them.

\begin{table}
    \caption{Static variables used for each of the datasets in the
    experiments. Categorical variables are shown in \textit{italics} and were
    expanded to one-hot encodings.} \label{tab:static variables}
    \begin{tabularx}{\linewidth}{p{0.25\linewidth}X}
        \toprule
        \textbf{Dataset} & \textbf{Static Variables} \\
        \midrule
        \dataset{M-Mortality} & Height \\
        \dataset{P-Mortality} & Age, \textit{Gender}, Height, \textit{ICUType} \\
        \dataset{P-Sepsis} & Age, \textit{Gender}, HospAdmTime \\
        \bottomrule
    \end{tabularx}
\end{table}

\paragraph{Time series variables}

For all datasets, we used all available time series variables including vitals,
lab measurements, and interventions. All variables were treated as
\emph{continuous}, and no additional transformations were applied.

\paragraph{Splits}

All datasets were partitioned into three subsets training, validation and
testing. For the \dataset{M-Mortality} dataset, the same splits as in
\cite{Harutyunyan2019} were used to ensure comparability of the obtained
results.  For both Physionet datasets~(\dataset{P-Mortality} and
\dataset{P-Sepsis}), we did not have access to the held-out test set used in
the challenges and thus defined our own splits. For this, the full dataset was
split into a training split~(80\%) and a testing split~(20\%), while
stratifying such that the splits have~(approximately) the same class imbalance.
This procedure was repeated on the training data to additionally create
a validation split. In the case of the online task \dataset{P-Sepsis},
stratification was based on whether the patient develops sepsis or not.

\paragraph{Implementation} We provide the complete data preprocessing
pipeline including the splits used to generate the results in this work
as a separate Python package \texttt{medical-ts-datasets}, which
integrates with \texttt{tensorflow-datasets}\citep{TFDS}.  This permits other
researchers to \emph{directly} compare to the results in this work. By
doing so, we strive to enable more rapid progress in the medical time
series community.

\subsection{Comparison partners}

The following paragraphs give a brief overview of the methods that we
used as comparison partners in our experiments.

\paragraph{GRU-simple}
\method{GRU-simple}~\citep{che2018recurrent} augments the input at time
$t$ of a Gated-Recurrent-Unit RNN with a measurement mask $m_t^d$ and a $\delta_t$
matrix, which contains the time since the last measurement of the corresponding
modality $d$, such that
\begin{equation*}
    \delta_t = \begin{cases}
        s_t - s_{t-1} + \delta_{t-1}^d & t > 1, m^d_{t-1} = 0 \\
        s_t - s_{t-1} & t > 1, m^d_{t-1} = 1 \\
        0             & t = 0
    \end{cases}
\end{equation*}
where $s_t$ represents the time associated with time step $t$.

\paragraph{GRU-D}
\method{GRU-D} or GRU-Decay \citep{che2018recurrent} contains modifications to the
GRU RNN cell, allowing it to decay past observations to the mean imputation of
a modality using a learnable decay rate. By additionally providing the
measurement masks as an input the recurrent neural network
the last feed in value. Learns how fast to decay back to a mean imputation of
the missing data modality.

\paragraph{Phased-LSTM}
The \method{Phased-LSTM} \citep{neil2016phased} introduced
a biologically inspired time dependent gating mechanism of a Long
short-term RNN cell \citep{hochreiter1997long}. This allows the network
to handle event-based sequences with irregularly spaced observations,
but not unaligned measurements. We thus additionally augment the input
in a similar fashion as described for the \method{GRU-simple} approach.

\paragraph{Interpolation Prediction Networks}
\method{IP-Networks} \citep{shukla2018interpolationprediction} apply
multiple semi-parametric interpolation schemes to irregularly-sampled
time series to obtain regularly-sampled representations that cover
long-term trends, transients, and also sampling information. The
parameters of the interpolation network are trained with the classifier
in an end-to-end fashion.

\paragraph{Transformer}
In the \method{Transformer} architecture \citep{vaswani2017attention},
the elements of a sequence are encoded simultaneously and information
between sequence elements is captured using Multi-Head-Attention blocks.
Transformers are typically used for sequence-to-sequence modelling
tasks. In our setup, we adapted them to classification tasks by
mean-aggregating the final representation.  This representation is then
fed into a one-layer MLP to predict logits for the individual classes.

\subsection{Implementation details}\label{appx:imp-details}

All experiments were run using \texttt{tensorflow 1.15.2} and training was
performed on \texttt{NVIDIA Geforce GTX 1080Ti} GPUs. In order to allow a fair
comparison between methods, the input processing pipeline employed
caching of model-specific representations and transformations of the data.

In contrast, due to the high complexity of the \method{Latent-ODE} model, we relied on the implementation provided by the authors and introduced our datasets into their code. This introduces the following differences between the evaluation of \method{Latent-ODE} compared to the other methods: \begin{inparaenum}
\item input processing pipeline is not cached
\item model code is written in PyTorch
\item due to an order of magnitude higher runtime, a thorough hyperparameter search was not feasible
\end{inparaenum}. This can introduce biases both in terms of runtime and
performance compared to the other methods.

\subsection{Training, Model Architectures, and Hyperparameter Search}
\label{appx:hyperparameter-search}

\paragraph{General}
All models were trained using the Adam optimizer~\citep{kingma2014adam},
while log-uniformly sampling the learning rate between $0.01$ and
$0.0001$. Further, the batch size of all methods was sampled from the
values $(32, 64, 128, 256, 512)$.

\paragraph{Recurrent neural networks}
For the RNN based methods (\method{GRU-Simple}, \method{Phased-LSTM},
\method{GRU-D} and \method{IP-Nets}), the number of units was sampled
from the values $(32, 64, 128, 256, 512, 1024)$. Further, recurrent dropout
and input dropout were sampled from the values $(0.0, 0.1, 0.2, 0.3, 0.4)$.
For the \method{Phased-LSTM} method, however, we did not apply dropout to the
recurrent state and the inputs, as the learnt frequencies were hypothesized to
fulfil a similar function as dropout \citep{neil2016phased}.  We
additionally sample parameters that are specific to \method{Phased-LSTM}: if
peephole connections should be used, the leak rate from $(0.001, 0.005, 0.01)$
and the maximal wavelength for initializing the hidden state phases from the
range $(10, 100, 1000)$. For \method{IP-Nets}, we additionally sample the
imputation stepsize uniformly from the range $(0.5, 1., 2.5, 5.)$ and the
fraction of reconstructed data points from $(0.05, 0.1, 0.2, 0.5,
0.75)$.

Static variables were handled by computing the initial hidden state of
the RNNs conditional on the static variables. For all methods, the
computation was performed using a one-hidden-layer neural network with
the number of hidden units set to the number of hidden units in the RNN. 

\paragraph{\methodname-Attn}
We vary the number of layers, dropout in between the layers and the number of
nodes per layer for both the encoding network $h_\theta$ and the aggregation
network $g_\psi$ from the same ranges. The number of layers is randomly sampled
between $1$ and $5$, the number of nodes in a layer are uniformly sampled from
the range $(16, 32, 64, 128, 256, 512)$ and the dropout fraction is sampled
from the values $(0.0, 0.1, 0.2, 0.3)$. The width of the embedding space prior
to aggregation is sampled from the values $(32, 64, 128, 256, 512, 1024,
2048)$. The aggregation function was set to be \texttt{sum} as described in the text.
The number of dimensions used for the positional embedding $\tau$ is selected
uniformly from $(4, 8, 16)$ and $\mathfrak{t}$, i.e.\ the maximum
time scale, was selected from the values $(10, 100, 1000)$.
The attention network $f'$ was set to always use \emph{mean} aggregation. 
Furthermore, we use a constant architecture for the attention network
$f'$ with $2$ layers, $64$ nodes per layer, 4 heads and a dimensionality
of the dot product space $d$ of $128$. We sample the amount of attention
dropout uniformly from the values $(0.0, 0.1, 0.25, 0.5)$.

\paragraph{Transformer}
We utilize the same model architecture as defined in
\citet{vaswani2017attention}, where we use an MLP with a single hidden
layer as a feed-forward network, with dimensionality of the hidden layer selected to be
twice the model dimensionality. The Transformer architecture was applied to the
time series by concatenating the vectors of each time point with a measurement
indicator. If no value was measured, input was set to zero for this modality.
The parameters for the Transformer network were sampled according to the
following criteria: the dimensionality of the model was sampled uniformly from
the values $(64, 128, 256, 512, 1024)$, the number of attention heads per layer
from the values $(2, 4, 8)$, and the number of layers from the range $[1,6] \in
\natural$. Moreover, we sampled the amount of dropout of the residual
connections and the amount of attention dropout uniformly from the values
$(0.0, 0.1, 0.2, 0.3, 0.5)$, and the maximal timescale for the time embedding
from the values $(10, 100, 1000)$~(similar to the \methodname approach).
Further, 1000 steps of warmup were applied, where the learning rate was
linearly scaled from $lr_{\min} = 0$ to the learning rate $lr_{\max}$ sampled by
the hyperparameter search.

\paragraph{Latent-ODE}
We utilize the implementation from \citet{rubanova2019latent} and extended the
evaluation metrics and datasets to fit our scenario.
Due to the long training time almost an order of magnitude longer than any
other method considered a thorough hyperparameter search as executed for the
other methods was not possible.  We thus rely on the hyperparameters selected
by the authors. In particular, we use their physionet 2012 dataset settings for
all datasets.
For further details see \prettyref{tab:selected hyperparameters}.

\paragraph{Selected hyperparameters}
In order to ensure reproducibility, the parameters selected by the
hyperparameter search are shown in \prettyref{tab:selected hyperparameters} for
all model dataset combinations.

\begin{table*}
    \caption{Best hyperparameters of all models on all datasets.} \label{tab:selected hyperparameters}
    \begin{tabularx}{\linewidth}{Xp{0.25\linewidth}p{0.25\linewidth}p{0.25\linewidth}}
\toprule
\textbf{Model} & \dataset{P-Mortality} & \dataset{M-Mortality} & \dataset{P-Sepsis} \\
\midrule
\method{GRU-D} & 
    lr: $0.00138$, bs: $512$, n\_units: $128$, dropout: $0.1$, recurrent\_dropout: $0.1$ &
    lr: $0.00016$, bs: $32$, n\_units: $256$, dropout: $0.0$, recurrent\_dropout: $0.2$ &
    lr: $0.0069$, bs: $128$, n\_units: $512$, dropout: $0.3$, recurrent\_dropout: $0.3$ \\
\midrule
\method{GRU-Simple} &
    lr: $0.00022$, bs: $256$, n\_units: $256$, dropout: $0.0$, recurrent\_dropout: $0.0$ &
    lr: $0.00011$, bs: $32$, n\_units: $512$, dropout: $0.3$, recurrent\_dropout: $0.4$ &
    lr: $0.00024$, bs: $64$, n\_units: $1024$, dropout: $0.3$, recurrent\_dropout: $0.3$ \\
\midrule
\method{IP-Nets}    &
    lr: $0.00035$, bs: $32$, n\_units: $32$, dropout: $0.4$, recurrent\_dropout: $0.3$, imputation\_stepsize: $1.0$, reconst\_fraction: $0.75$ &
    lr: $0.00062$, bs: $16$, n\_units: $256$, dropout: $0.2$, recurrent\_dropout: $0.1$, imputation\_stepsize: $1.0$, reconst\_fraction: $0.2$ &
    lr: $0.0008$, bs: $16$, n\_units: $32$, dropout: $0.3$, recurrent\_dropout: $0.4$, imputation\_stepsize: $1.0$, reconst\_fraction: $0.5$ \\
\midrule
\method{Transformer} & 
    lr: $0.00567$, bs: $256$, warmup\_steps: $1000$, n\_dims: $512$, n\_heads: $2$, n\_layers: $1$, dropout: $0.3$, attn\_dropout: $0.3$, aggregation\_fn: max, max\_timescale: $1000.0$ &
    lr: $0.00204$, bs: $256$, warmup\_steps: $1000$, n\_dims: $512$, n\_heads: $8$, n\_layers: $2$, dropout: $0.4$, attn\_dropout: $0.0$, aggregation\_fn: mean, max\_timescale: $100.0$ &
    lr: $0.00027$, bs: $128$, warmup\_steps: $1000$, n\_dims: $128$, n\_heads: $2$, n\_layers: $4$, dropout: $0.1$, attn\_dropout: $0.4$, aggregation\_fn: mean, max\_timescale: $100.0$ \\
\midrule
\method{Phased-LSTM} &
    lr: $0.00262$, bs: $256$, n\_units: $128$, use\_peepholes: True, leak: $0.01$, period\_init\_max: $1000.0$ &
    lr: $0.00576$, bs: $32$, n\_units: $1024$, use\_peepholes: False, leak: $0.01$, period\_init\_max: $1000.0$ &
    lr: $0.00069$, bs: $32$, n\_units: $512$, use\_peepholes: False, leak: $0.001$, period\_init\_max: $100.0$ \\
\midrule
\method{Latent-ODE} & 
    optimizer: Adamax, lr\_schedule: exponential decay, lr: $0.01$, bs: $50$, rec-dims: $40$, rec-layers: $3$ gen-layers: $3$, units: $50$, gru-units: $50$, quantization: $0.016$, classification: True, reconstruction: True &
    optimizer: Adamax, lr\_schedule: exponential decay, lr: $0.01$, bs: $50$, rec-dims: $40$, rec-layers: $3$ gen-layers: $3$, units: $50$, gru-units: $50$, quantization: $0.016$, classification: True, reconstruction: True &
    optimizer: Adamax, lr\_schedule: exponential decay, lr: $0.01$, bs: $50$, rec-dims: $40$, rec-layers: $3$, gen-layers: $3$, units: $50$, gru-units: $50$, quantization: $1$, classification: True, reconstruction: True \\
\midrule
\method{\methodname-Attn} &
    lr: $0.00081$, bs: $512$, n\_phi\_layers: $4$, phi\_width: $128$, phi\_dropout: $0.2$, n\_psi\_layers: $2$, psi\_width: $64$, psi\_latent\_width: $128$, dot\_prod\_dim: $128$, n\_heads: $4$, attn\_dropout: $0.5$, latent\_width: $32$, n\_rho\_layers: $2$, rho\_width: $512$, rho\_dropout: $0.0$, max\_timescale: $100.0$, n\_positional\_dims: $4$ &
    lr: $0.00245$, bs: $512$, n\_phi\_layers: $3$, phi\_width: $64$, phi\_dropout: $0.1$, n\_psi\_layers: $2$, psi\_width: $64$, psi\_latent\_width: $128$, dot\_prod\_dim: $128$, n\_heads: $4$, attn\_dropout: $0.1$, latent\_width: $256$, n\_rho\_layers: $2$, rho\_width: $512$, rho\_dropout: $0.1$, max\_timescale: $1000.0$, n\_positional\_dims: $8$ &
    lr: $0.00011$, bs: $64$, n\_phi\_layers: $4$, phi\_width: $32$, phi\_dropout: $0.0$, n\_psi\_layers: $2$, psi\_width: $64$, psi\_latent\_width: $128$, dot\_prod\_dim: $128$, n\_heads: $4$, attn\_dropout: $0.1$, latent\_width: $512$, n\_rho\_layers: $3$, rho\_width: $128$, rho\_dropout: $0.0$, max\_timescale: $10.0$, n\_positional\_dims: $16$ \\
\bottomrule
\end{tabularx}
\end{table*}

\end{document}

%% file: tables/the_big_boss.tex
\newcommand{\centeredOOM}{\phantom{X}---\OOM---\phantom{X}}
\renewrobustcmd{\bfseries}{\fontseries{b}\selectfont}
\renewrobustcmd{\boldmath}{}
\setlength{\tabcolsep}{2.5pt}
\sisetup{%
  table-format            = 2.2,
  detect-weight           = true,
  detect-shape            = true,
  detect-all              = true,
  mode                    = text,
}
{%
\scriptsize
\begin{tabular}{
  @{}
  l
  l
  S
  S
  S
  S
  @{}
}
\toprule
    Dataset & Model                     & \multicolumn{1}{c}{Accuracy}  & \multicolumn{1}{c}{AUPRC}     & \multicolumn{1}{c}{AUROC}     & \multicolumn{1}{c}{\nicefrac{\si{\second}\,}{epoch}} \\
\midrule
    \multirow{6}{*}{\dataset{M3M}}
        & \method{GRU-D}                &          \num{77.0 \pm 1.5} & \bfseries\num{52.0 \pm 0.8} & \bfseries\num{85.7 \pm 0.2} &          \num{133 \pm 8} \\
        & \method{GRU-Simple}           &          \num{78.1 \pm 1.3} &          \num{43.6 \pm 0.4} &          \num{82.8 \pm 0.0} &          \num{140 \pm 7} \\
        & \method{IP-Nets}              &          \num{78.3 \pm 0.7} & \itshape \num{48.3 \pm 0.4} &          \num{83.2 \pm 0.5} &           \num{81.2 \pm 8.5} \\
        & \method{Phased-LSTM}          &          \num{73.8 \pm 3.3} &          \num{37.1 \pm 0.5} &          \num{80.3 \pm 0.4} &          \num{166 \pm 7} \\
        & \method{Transformer}          & \itshape \num{77.4 \pm 5.6} &          \num{42.6 \pm 1.0} &          \num{82.1 \pm 0.3} & \itshape  \num{20.1 \pm 0.1} \\
        & \method{Latent-ODE}$^\dagger$ &          \num{72.8 \pm 1.7} &          \num{39.5 \pm 0.5} &          \num{80.9 \pm 0.2} &         \num{4622} \\
        & \method{\methodname-Attn}     & \bfseries\num{79.0 \pm 2.2} &          \num{46.3 \pm 0.5} & \itshape \num{83.9 \pm 0.4} & \bfseries \num{14.5 \pm 0.5} \\
\midrule
    \multirow{6}{*}{\dataset{P12}}
        & \method{GRU-D}                &          \num{80.0 \pm 2.9} & \itshape \num{53.7 \pm 0.9} & \bfseries\num{86.3 \pm 0.3} &            \num{8.67 \pm 0.49} \\
        & \method{GRU-Simple}           & \itshape \num{82.2 \pm 0.2} &          \num{42.2 \pm 0.6} &          \num{80.8 \pm 1.1} &           \num{30.0 \pm 2.5} \\
        & \method{IP-Nets}              &          \num{79.4 \pm 0.3} &          \num{51.0 \pm 0.6} & \itshape \num{86.0 \pm 0.2} &           \num{25.3 \pm 1.8} \\
        & \method{Phased-LSTM}          &          \num{76.8 \pm 5.2} &          \num{38.7 \pm 1.5} &          \num{79.0 \pm 1.0} &           \num{44.6 \pm 2.3} \\
        & \method{Transformer}          & \bfseries\num{83.7 \pm 3.5} & \bfseries\num{52.8 \pm 2.2} & \bfseries\num{86.3 \pm 0.8} & \bfseries  \num{6.06 \pm 0.06} \\
        & \method{Latent-ODE}$^\dagger$ &          \num{76.0 \pm 0.1} &          \num{50.7 \pm 1.7} &          \num{85.7 \pm 0.6} &         \num{3500} \\
        & \method{\methodname-Attn}     &          \num{75.3 \pm 3.5} &          \num{52.4 \pm 1.1} &          \num{85.1 \pm 0.4} & \itshape   \num{7.62 \pm 0.10} \\
\bottomrule
\end{tabular}
}

%% file: tables/online.tex
\setlength{\tabcolsep}{2.5pt}
\newcommand{\leakage}[1]{\color{gray}#1}
\sisetup{%
  table-format            = 2.2,
  detect-weight           = true,
  detect-shape            = true,
  detect-all              = true,
  mode                    = text,
}
{%
\scriptsize %
\begin{tabular}{
  @{}
  l
  S
  S
  S
  S
  S
  @{}
}
\toprule
    Model                         & \multicolumn{1}{c}{B-Accuracy} & \multicolumn{1}{c}{AUPRC}  & \multicolumn{1}{c}{AUROC}     & \multicolumn{1}{c}{$U_{\text{norm}}$} & \multicolumn{1}{c}{\nicefrac{\si{\second}\,}{epoch}} \\
\midrule
    \method{GRU-D}                &          \num{57.4 \pm 0.2} &           \num{5.33 \pm 0.39} &          \num{67.4 \pm 1.2} &          \num{12.6 \pm 1.1} & \itshape \num{72.3} \\
    \method{GRU-Simple}           & \bfseries\num{71.0 \pm 1.4} & \bfseries \num{6.10 \pm 0.75} & \bfseries\num{78.1 \pm 1.5} & \bfseries\num{26.9 \pm 4.1} &         \num{116} \\
    \method{IP-Nets}              & \leakage{\num{87.1 \pm 0.9}}& \leakage{\num{29.4 \pm 2.1}}  & \leakage{\num{94.1 \pm 0.4}}& \leakage{\num{62.2 \pm 1.3}}&         \num{253} \\
    \method{IP-Nets}$^*$          &          \num{63.8 \pm 0.9} &           \num{5.11 \pm 0.80} &          \num{74.2 \pm 1.2} &         \num{-11.9 \pm 4.0} &         \num{253} \\
    \method{Phased-LSTM}          &          \num{67.5 \pm 1.7} & \itshape  \num{5.54 \pm 0.91} &          \num{75.4 \pm 1.3} &          \num{20.2 \pm 3.2} &         \num{192} \\
    \method{Latent-ODE}$^\dagger$ &          \num{62.4 \pm 0.1} &          \num{11.4 \pm 2.1}   &          \num{64.6 \pm 0.7} &          \num{12.3 \pm 1.0} &         \num{1872} \\
    \method{Transformer}          & \leakage{\num{91.2 \pm 0.2}}& \leakage{\num{53.4 \pm 5.6}}  & \leakage{\num{97.3 \pm 0.2}}& \leakage{\num{71.3 \pm 1.4}}& \bfseries\num{28.5} \\
    \method{Transformer}$^*$      &          \num{53.6 \pm 1.7} &           \num{3.63 \pm 0.95} &          \num{65.8 \pm 3.7} &         \num{-43.9 \pm 10.0}& \bfseries\num{28.5} \\
    \method{\methodname-Attn}     & \itshape \num{70.9 \pm 0.8} &           \num{4.84 \pm 0.22} & \itshape \num{76.8 \pm 0.9} & \itshape \num{25.6 \pm 1.9} &          \num{77.5} \\
\bottomrule
\end{tabular}
}

%% file: tables/mimic_binary_prevalences.tex
\begin{tabularx}{\linewidth}{llll}
 \toprule
  & Train & Test & Val\\ \midrule
  In-hospital deaths & \num{0.135} & \num{0.116} & \num{0.135}  \\
\bottomrule
\end{tabularx}

%% file: tables/physionet_binary_prevalences.tex
\begin{tabularx}{\linewidth}{llll}
 \toprule
  & Train & Test & Val\\ \midrule
  In-hospital deaths & \num{0.142} & \num{0.142} & \num{0.142}\\
\bottomrule
\end{tabularx}

%% file: main.bbl
\begin{thebibliography}{44}
\providecommand{\natexlab}[1]{#1}
\providecommand{\url}[1]{\texttt{#1}}
\expandafter\ifx\csname urlstyle\endcsname\relax
  \providecommand{\doi}[1]{doi: #1}\else
  \providecommand{\doi}{doi: \begingroup \urlstyle{rm}\Url}\fi

\bibitem[TFD()]{TFDS}
{TensorFlow Datasets}, a collection of ready-to-use datasets.
\newblock \url{https://www.tensorflow.org/datasets}.

\bibitem[Bachlechner et~al.(2020)Bachlechner, Majumder, Mao, Cottrell, and
  McAuley]{bachlechner2020rezero}
Bachlechner, T., Majumder, B.~P., Mao, H.~H., Cottrell, G.~W., and McAuley, J.
\newblock Rezero is all you need: Fast convergence at large depth.
\newblock \emph{arXiv preprint arXiv:2003.04887}, 2020.

\bibitem[Bahadori \& Lipton(2019)Bahadori and Lipton]{Bahadori19}
Bahadori, M.~T. and Lipton, Z.~C.
\newblock Temporal-clustering invariance in irregular healthcare time series.
\newblock \emph{arXiv preprint arXiv:1904.12206}, 2019.

\bibitem[Bonilla et~al.(2008)Bonilla, Chai, and Williams]{bonilla2008multi}
Bonilla, E.~V., Chai, K.~M., and Williams, C.
\newblock Multi-task gaussian process prediction.
\newblock In \emph{Advances in Neural Information Processing
  Systems~(NeurIPS)}, pp.\  153--160, 2008.

\bibitem[Che et~al.(2018)Che, Purushotham, Cho, Sontag, and
  Liu]{che2018recurrent}
Che, Z., Purushotham, S., Cho, K., Sontag, D., and Liu, Y.
\newblock Recurrent neural networks for multivariate time series with missing
  values.
\newblock \emph{Scientific reports}, 8\penalty0 (1):\penalty0 6085, 2018.

\bibitem[Desautels et~al.(2016)Desautels, Calvert, Hoffman, Jay, Kerem, Shieh,
  Shimabukuro, Chettipally, Feldman, Barton, et~al.]{desautels2016prediction}
Desautels, T., Calvert, J., Hoffman, J., Jay, M., Kerem, Y., Shieh, L.,
  Shimabukuro, D., Chettipally, U., Feldman, M.~D., Barton, C., et~al.
\newblock Prediction of sepsis in the intensive care unit with minimal
  electronic health record data: {A} machine learning approach.
\newblock \emph{JMIR Medical Informatics}, 4\penalty0 (3):\penalty0 e28, 2016.

\bibitem[Futoma et~al.(2017)Futoma, Hariharan, and Heller]{futoma2017learning}
Futoma, J., Hariharan, S., and Heller, K.
\newblock Learning to detect sepsis with a {M}ultitask {G}aussian {P}rocess
  {RNN} classifier.
\newblock In \emph{International Conference on Machine Learning~(ICML)}, pp.\
  1174--1182, 2017.

\bibitem[Garnelo et~al.(2018)Garnelo, Rosenbaum, Maddison, Ramalho, Saxton,
  Shanahan, Teh, Rezende, and Eslami]{garnelo2018conditional}
Garnelo, M., Rosenbaum, D., Maddison, C., Ramalho, T., Saxton, D., Shanahan,
  M., Teh, Y.~W., Rezende, D., and Eslami, S.~A.
\newblock Conditional neural processes.
\newblock In \emph{International Conference on Machine Learning}, pp.\
  1690--1699, 2018.

\bibitem[Goldberger et~al.(2000)Goldberger, Amaral, Glass, Hausdorff, Ivanov,
  Mark, Mietus, Moody, Peng, and Stanley]{goldberger2000physiobank}
Goldberger, A.~L., Amaral, L.~A., Glass, L., Hausdorff, J.~M., Ivanov, P.,
  Mark, R.~G., Mietus, J.~E., Moody, G.~B., Peng, C.-K., and Stanley, H.~E.
\newblock Physiobank, physiotoolkit, and physionet: components of a new
  research resource for complex physiologic signals.
\newblock \emph{Circulation}, 101\penalty0 (23):\penalty0 e215--e220, 2000.

\bibitem[Harutyunyan et~al.(2019)Harutyunyan, Khachatrian, Kale, Ver~Steeg, and
  Galstyan]{Harutyunyan2019}
Harutyunyan, H., Khachatrian, H., Kale, D.~C., Ver~Steeg, G., and Galstyan, A.
\newblock Multitask learning and benchmarking with clinical time series data.
\newblock \emph{Scientific Data}, 6\penalty0 (1):\penalty0 96, 2019.
\newblock ISSN 2052-4463.

\bibitem[Hawkes(1971)]{Hawkes71}
Hawkes, A.~G.
\newblock Spectra of some self-exciting and mutually exciting point processes.
\newblock \emph{Biometrika}, 58\penalty0 (1):\penalty0 83--90, 1971.

\bibitem[Hochreiter \& Schmidhuber(1997)Hochreiter and
  Schmidhuber]{hochreiter1997long}
Hochreiter, S. and Schmidhuber, J.
\newblock Long short-term memory.
\newblock \emph{Neural Computation}, 9\penalty0 (8):\penalty0 1735--1780, 1997.

\bibitem[Johnson et~al.(2016)Johnson, Pollard, Shen, Lehman, Feng, Ghassemi,
  Moody, Szolovits, Anthony~Celi, and Mark]{johnson2016mimic}
Johnson, A. E.~W., Pollard, T.~J., Shen, L., Lehman, L.-w.~H., Feng, M.,
  Ghassemi, M., Moody, B., Szolovits, P., Anthony~Celi, L., and Mark, R.~G.
\newblock {MIMIC-III}, a freely accessible critical care database.
\newblock \emph{Scientific Data}, 3, 2016.

\bibitem[Kahan(1965)]{Kahan65}
Kahan, W.
\newblock Pracniques: Further remarks on reducing truncation errors.
\newblock \emph{Communications of the ACM}, 8\penalty0 (1):\penalty0 40--41,
  1965.

\bibitem[Kingma \& Ba(2015)Kingma and Ba]{kingma2014adam}
Kingma, D.~P. and Ba, J.
\newblock Adam: A method for stochastic optimization.
\newblock In \emph{International Conference on Learning Representations}, 2015.

\bibitem[Knuth(1998)]{Knuth98}
Knuth, D.
\newblock \emph{The Art of Computer Programming, Volume 2}.
\newblock Addison-Wesley, 1998.

\bibitem[Lee et~al.(2019)Lee, Lee, Kim, Kosiorek, Choi, and Teh]{lee2019set}
Lee, J., Lee, Y., Kim, J., Kosiorek, A., Choi, S., and Teh, Y.~W.
\newblock Set transformer: A framework for attention-based
  permutation-invariant neural networks.
\newblock In Chaudhuri, K. and Salakhutdinov, R. (eds.), \emph{Proceedings of
  the 36th International Conference on Machine Learning}, volume~97 of
  \emph{Proceedings of Machine Learning Research}, pp.\  3744--3753, Long
  Beach, California, USA, 09--15 Jun 2019. PMLR.

\bibitem[Li \& Marlin(2015)Li and Marlin]{li2015classification}
Li, S. C.-X. and Marlin, B.~M.
\newblock Classification of sparse and irregularly sampled time series with
  mixtures of expected gaussian kernels and random features.
\newblock In \emph{UAI}, pp.\  484--493, 2015.

\bibitem[Li \& Marlin(2016)Li and Marlin]{li2016scalable}
Li, S. C.-X. and Marlin, B.~M.
\newblock A scalable end-to-end {G}aussian process adapter for irregularly
  sampled time series classification.
\newblock In \emph{Advances In Neural Information Processing Systems~29}, pp.\
  1804--1812, 2016.

\bibitem[Liebovitch \& Toth(1989)Liebovitch and Toth]{Liebovitch89}
Liebovitch, L.~S. and Toth, T.
\newblock A fast algorithm to determine fractal dimensions by box counting.
\newblock \emph{Physics Letters A}, 141\penalty0 (8):\penalty0 386--390, 1989.

\bibitem[Liniger(2009)]{Liniger09}
Liniger, T.~J.
\newblock \emph{Multivariate {H}awkes processes}.
\newblock PhD thesis, ETH Zurich, 2009.

\bibitem[Lipton et~al.(2016)Lipton, Kale, and Wetzel]{lipton2016directly}
Lipton, Z.~C., Kale, D., and Wetzel, R.
\newblock Directly modeling missing data in sequences with rnns: Improved
  classification of clinical time series.
\newblock In \emph{Machine Learning for Healthcare Conference}, pp.\  253--270,
  2016.

\bibitem[Little \& Rubin(2014)Little and Rubin]{little2014statistical}
Little, R.~J. and Rubin, D.~B.
\newblock \emph{Statistical analysis with missing data}, volume 333.
\newblock John Wiley \& Sons, 2014.

\bibitem[Lu et~al.(2008)Lu, Leen, Huang, and Erdogmus]{lu2008reproducing}
Lu, Z., Leen, T.~K., Huang, Y., and Erdogmus, D.
\newblock A reproducing kernel hilbert space framework for pairwise time series
  distances.
\newblock In \emph{Proceedings of the 25th International Conference on Machine
  learning}, pp.\  624--631, 2008.

\bibitem[Lukasik et~al.(2016)Lukasik, Srijith, Vu, Bontcheva, Zubiaga, and
  Cohn]{Lukasik16}
Lukasik, M., Srijith, P.~K., Vu, D., Bontcheva, K., Zubiaga, A., and Cohn, T.
\newblock {H}awkes processes for continuous time sequence classification: an
  application to rumour stance classification in {T}witter.
\newblock In \emph{Proceedings of the 54th Annual Meeting of the Association
  for Computational Linguistics (Volume 2: Short Papers)}, pp.\  393--398,
  Berlin, Germany, August 2016. Association for Computational Linguistics.
\newblock \doi{10.18653/v1/P16-2064}.
\newblock URL \url{https://www.aclweb.org/anthology/P16-2064}.

\bibitem[Marlin et~al.(2012)Marlin, Kale, Khemani, and
  Wetzel]{marlin2012unsupervised}
Marlin, B.~M., Kale, D.~C., Khemani, R.~G., and Wetzel, R.~C.
\newblock {Unsupervised pattern discovery in electronic health care data using
  probabilistic clustering models}.
\newblock In \emph{Proceedings of the 2nd ACM SIGHIT International Health
  Informatics Symposium}, pp.\  389--398. ACM, 2012.

\bibitem[Mei \& Eisner(2017)Mei and Eisner]{Mei17}
Mei, H. and Eisner, J.~M.
\newblock The neural {H}awkes process: A neurally self-modulating multivariate
  point process.
\newblock In Guyon, I., Luxburg, U.~V., Bengio, S., Wallach, H., Fergus, R.,
  Vishwanathan, S., and Garnett, R. (eds.), \emph{Advances in Neural
  Information Processing Systems~30}, pp.\  6754--6764. Curran Associates,
  Inc., 2017.

\bibitem[Moor et~al.(2019)Moor, Horn, Rieck, Roqueiro, and Borgwardt]{Moor19}
Moor, M., Horn, M., Rieck, B., Roqueiro, D., and Borgwardt, K.
\newblock Early recognition of sepsis with gaussian process temporal
  convolutional networks and dynamic time warping.
\newblock In Doshi-Velez, F., Fackler, J., Jung, K., Kale, D., Ranganath, R.,
  Wallace, B., and Wiens, J. (eds.), \emph{Proceedings of the 4th Machine
  Learning for Healthcare Conference}, volume 106 of \emph{Proceedings of
  Machine Learning Research}, pp.\  2--26, Ann Arbor, Michigan, 09--10 Aug
  2019. PMLR.

\bibitem[Neil et~al.(2016)Neil, Pfeiffer, and Liu]{neil2016phased}
Neil, D., Pfeiffer, M., and Liu, S.-C.
\newblock Phased lstm: Accelerating recurrent network training for long or
  event-based sequences.
\newblock In \emph{Advances in Neural Information Processing Systems~29}, pp.\
  3882--3890, 2016.

\bibitem[Nguyen \& Salazar(2019)Nguyen and Salazar]{nguyen2019transformers}
Nguyen, T.~Q. and Salazar, J.
\newblock Transformers without tears: Improving the normalization of
  self-attention.
\newblock \emph{arXiv preprint arXiv:1910.05895}, 2019.

\bibitem[Reyna et~al.(2020)Reyna, Josef, Jeter, Shashikumar, Westover, Nemati,
  Clifford, and Sharma]{reyna2020early}
Reyna, M.~A., Josef, C.~S., Jeter, R., Shashikumar, S.~P., Westover, M.~B.,
  Nemati, S., Clifford, G.~D., and Sharma, A.
\newblock Early prediction of sepsis from clinical data: the
  physionet/computing in cardiology challenge 2019.
\newblock \emph{Critical care medicine}, 48\penalty0 (2):\penalty0 210--217,
  2020.

\bibitem[Rubanova et~al.(2019)Rubanova, Chen, and Duvenaud]{rubanova2019latent}
Rubanova, Y., Chen, R.~T., and Duvenaud, D.
\newblock Latent odes for irregularly-sampled time series.
\newblock \emph{arXiv preprint arXiv:1907.03907}, 2019.

\bibitem[Seymour et~al.(2016)Seymour, Liu, Iwashyna, Brunkhorst, Rea, Scherag,
  Rubenfeld, Kahn, Shankar-Hari, Singer, et~al.]{seymour2016assessment}
Seymour, C.~W., Liu, V.~X., Iwashyna, T.~J., Brunkhorst, F.~M., Rea, T.~D.,
  Scherag, A., Rubenfeld, G., Kahn, J.~M., Shankar-Hari, M., Singer, M., et~al.
\newblock Assessment of clinical criteria for sepsis: for the third
  international consensus definitions for sepsis and septic shock (sepsis-3).
\newblock \emph{Jama}, 315\penalty0 (8):\penalty0 762--774, 2016.

\bibitem[Shukla \& Marlin(2019)Shukla and
  Marlin]{shukla2018interpolationprediction}
Shukla, S.~N. and Marlin, B.
\newblock Interpolation-prediction networks for irregularly sampled time
  series.
\newblock In \emph{International Conference on Learning Representations}, 2019.

\bibitem[Song et~al.(2018)Song, Rajan, Thiagarajan, and
  Spanias]{song2018attend}
Song, H., Rajan, D., Thiagarajan, J.~J., and Spanias, A.
\newblock Attend and diagnose: Clinical time series analysis using attention
  models.
\newblock In \emph{Thirty-second AAAI conference on artificial intelligence},
  2018.

\bibitem[Takens(1981)]{Takens81}
Takens, F.
\newblock Detecting strange attractors in turbulence.
\newblock In Rand, D. and Young, L.-S. (eds.), \emph{Dynamical Systems and
  Turbulence}, pp.\  366--381, Heidelberg, Germany, 1981. Springer.

\bibitem[Vaswani et~al.(2017)Vaswani, Shazeer, Parmar, Uszkoreit, Jones, Gomez,
  Kaiser, and Polosukhin]{vaswani2017attention}
Vaswani, A., Shazeer, N., Parmar, N., Uszkoreit, J., Jones, L., Gomez, A.~N.,
  Kaiser, {\L}., and Polosukhin, I.
\newblock Attention is all you need.
\newblock In \emph{Advances in Neural Information Processing
  Systems~(NeurIPS)}, pp.\  5998--6008, 2017.

\bibitem[Vinyals et~al.(2016)Vinyals, Bengio, and Kudlur]{vinyals2015order}
Vinyals, O., Bengio, S., and Kudlur, M.
\newblock Order matters: Sequence to sequence for sets.
\newblock In \emph{International Conference on Learning Representations}, 2016.

\bibitem[Wagstaff et~al.(2019)Wagstaff, Fuchs, Engelcke, Posner, and
  Osborne]{Wagstaff2019}
Wagstaff, E., Fuchs, F.~B., Engelcke, M., Posner, I., and Osborne, M.
\newblock On the limitations of representing functions on sets.
\newblock \emph{arXiv preprint arXiv:1901.09006}, 2019.

\bibitem[Williams \& Rasmussen(2006)Williams and
  Rasmussen]{williams2006gaussian}
Williams, C.~K. and Rasmussen, C.~E.
\newblock Gaussian processes for machine learning.
\newblock \emph{MIT Press}, 2\penalty0 (3):\penalty0 4, 2006.

\bibitem[Xiao et~al.(2017)Xiao, Yan, Farajtabar, Song, Yang, and Zha]{Xiao17}
Xiao, S., Yan, J., Farajtabar, M., Song, L., Yang, X., and Zha, H.
\newblock Joint modeling of event sequence and time series with attentional
  twin recurrent neural networks.
\newblock \emph{arXiv preprint arXiv:1703.08524}, 2017.

\bibitem[Yadav et~al.(2018)Yadav, Steinbach, Kumar, and Simon]{yadav2018mining}
Yadav, P., Steinbach, M., Kumar, V., and Simon, G.
\newblock Mining electronic health records ({EHRs}): a survey.
\newblock \emph{ACM Computing Surveys}, 50\penalty0 (6):\penalty0 85, 2018.

\bibitem[Yang et~al.(2017)Yang, Etesami, He, and Kiyavash]{Yang17}
Yang, Y., Etesami, J., He, N., and Kiyavash, N.
\newblock Online learning for multivariate hawkes processes.
\newblock In Guyon, I., Luxburg, U.~V., Bengio, S., Wallach, H., Fergus, R.,
  Vishwanathan, S., and Garnett, R. (eds.), \emph{Advances in Neural
  Information Processing Systems~30}, pp.\  4937--4946. Curran Associates,
  Inc., 2017.

\bibitem[Zaheer et~al.(2017)Zaheer, Kottur, Ravanbakhsh, Poczos, Salakhutdinov,
  and Smola]{zaheer2017deep}
Zaheer, M., Kottur, S., Ravanbakhsh, S., Poczos, B., Salakhutdinov, R.~R., and
  Smola, A.~J.
\newblock Deep sets.
\newblock In \emph{Advances in Neural Information Processing
  Systems~(NeurIPS)}, pp.\  3391--3401, 2017.

\end{thebibliography}
